\documentclass[11pt,a4paper]{article}
\usepackage[hyperref]{acl2017}
\usepackage{times}
\usepackage{latexsym}

\usepackage{url}

\usepackage{graphicx}
\usepackage{array}
\usepackage{multirow}

\usepackage{amsmath}
\usepackage{changepage}

\aclfinalcopy 

\title{Extracting Bilingual Persian Italian Lexicon from Comparable Corpora Using Different Types of Seed Dictionaries}

\author{Ebrahim Ansari,\footnotemark[2] \footnotemark[3] ~  M.H. Sadreddini,\footnotemark[4] ~
Lucio Grandinetti,\footnotemark[5] ~ 
\\
\textbf{Mahsa Radinmehr,\footnotemark[2] ~ Ziba Khosravan,\footnotemark[2] ~ and Mehdi Sheikhalishahi\footnotemark[2]}\\
\footnotemark[2] ~Department of Computer Science and Information Technology,\\ Institute for Advanced Studies in Basic Sciences\\
\footnotemark[3] ~Institute of Formal and Applied Linguistics,\\ Faculty of Mathematics and Physics, Charles University\\
\footnotemark[4] ~Computer Science and Engineering Department, Shiraz University\\
\footnotemark[5] ~Department of Electronics, Informatics and Systems, University of Calabria \\
  {\tt ansari@iasbs.ac.ir}
}
\date{}

\begin{document}
\maketitle
\begin{abstract}
Bilingual dictionaries are very important in various fields of natural language processing. In recent years, research on extracting new bilingual lexicons from non-parallel (comparable) corpora have been proposed. Almost all use a small existing dictionary or other resources to make an initial list called the ``seed dictionary". In this paper, we discuss the use of different types of dictionaries as the initial starting list for creating a bilingual Persian-Italian lexicon from a comparable corpus. Our experiments apply state-of-the-art techniques on three different seed dictionaries; an existing dictionary, a dictionary created with pivot-based schema, and a dictionary extracted from a small Persian-Italian parallel text. The interesting challenge of our approach is to find a way to combine different dictionaries together in order to produce a better and more accurate lexicon. In order to combine seed dictionaries, we propose two different combination models and examine the effect of our novel combination models on various comparable corpora that have differing degrees of comparability. We conclude with a proposal for a new weighting system to improve the extracted lexicon. The experimental results produced by our implementation show the efficiency of our proposed models.\footnote{Accepted to be published in ”Applications of Comparable Corpora”, Berlin:
Language Science Press.}
\end{abstract}

\section{Introduction}
\label{section1}
Bilingual lexicons are a key resource in a multilingual society. Their application can be found in a range of activities such as translation, language learning, or as a basic resource for natural language processing (NLP). The availability of translation resources varies depending on the language pairs. Therefore, bilingual dictionaries for languages with fewer native speakers are scarce or even non-existent. There are many papers describing different methods for building bilingual dictionaries automatically. Though automatic methods often have drawbacks such as including noise in the form of erroneous translations of some words, they are still popular because the alternative -- manually constructing a dictionary -- is very time-consuming. Automatic methods are often used to generate a first noisy dictionary that can be cleaned up and extended by manual work \citep{sjobergh_creating_2005}.

A pivot language (bridge language) is useful for creating bilingual resources such as bilingual dictionaries. The Pivot-based bilingual dictionary building is based on merging two bilingual dictionaries that share a common language. For example, using the Persian-English and the English-Italian dictionaries to build a new Persian-Italian lexicon. In recent years, some approaches based on this idea have been proposed \citep{tanaka_construction_1994, sjobergh_creating_2005, istvan_bilingual_2009, tsunakawa_building_2008, tsunakawa_improving_2013, ahn_automatic_2006}. The main problem of these methods is the amount of noise in extracted dictionaries when they have many incorrect translations.

In the last decade, some research has been proposed to acquire bilingual lexicons from non-parallel (comparable) corpora. A comparable corpus consists of sets of documents in several languages dealing with a given topic, or domain when documents have been composed independently of each other in different languages. Unlike the parallel corpora, which are clearly defined as translated text, there is a wide variation of non-parallelism in comparable texts \citep{Ansari2014b}. In comparison with parallel corpus, comparable corpora are much easier to build from commonly available documents, such as news article pairs describing the same event in different languages. Therefore, there is a growing interest in acquiring bilingual lexicons from comparable corpora. These methods have been proposed to extract a lexicon from comparable corpora when a suitable lexicon does not exist or is not complete enough. These methods are based on this assumption: there is a correlation between co-occurrence patterns in different languages \citep{rapp_identifying_1995}. For example, if the words teacher and school co-occur more often than expected by chance in an English corpus then the German translations of teacher and school, \textit{Lehrer} and \textit{schule}, should also co-occur more often than expected in a German corpus \citep{rapp_identifying_1995}.

In recent years, many methods have been proposed to build bilingual dictionaries based on the above correlation. Most of them share a standard strategy based on context similarity. The basis of these methods is finding the target words that have the most similar distributions with a given source word. The starting point of this strategy is a list of bilingual expressions that are used to build the context vectors of all words in both languages. This starting list, or initial dictionary, is named the seed dictionary \citep{fung_compiling_1995} and is usually provided by an external bilingual dictionary \citep{rapp_automatic_1999, chiao_looking_2002, fung_finding_1997, fung_ir_1998}. Some of the recent methods use small parallel corpora to create their seed list \citep{otero_learning_2007} and some uses no dictionary for starting phases \citep{rapp_utilizing_2010}. Sometimes there are different types of dictionaries, with each having its own accuracy. \citep{Ansari2014a} propose two simple methods to combine four different dictionaries (one existing dictionary and three dictionaries extracted using pivot based method) to increase the accuracy of the output. They use three languages English, Arabic, and French to create their pivot based lexicons.

In this work, we use three different types of dictionaries and then combine them to create our seed dictionaries. The first dictionary is a small existing Persian-Italian dictionary. The second dictionary is extracted from a pivot-based method. The third dictionary is created from our small parallel Persian-Italian corpus. Using these dictionaries, we propose different model combinations and a new weighting method to use on these different dictionaries. In comparison with \citep{Ansari2014a}'s approach, we introduce some new combination schemas to improve the quality of the result seed dictionary. Moreover, a parallel based extracted lexicon also is used as one of our initial seed dictionaries. Contrary to previous work, we apply our idea on various comparable corpora that have different degrees of comparability.

Section \ref{section2} reviews our approach and Section \ref{section3} describes our approach. Section \ref{section4} describes the methodology and resources used in our work; Section \ref{section5} shows the experimental results and finally, Section \ref{section6} concludes our work.

\section{Related works}
\label{section2}
In this Section, we discuss approaches and implementations in three parts and show their relation to our work. Section \ref{subsection21} describes the process of building a bilingual lexicon by using a pivot language using source-pivot and pivot-target dictionaries. In Section \ref{subsection22}, the idea of using parallel corpora to extract a bilingual dictionary is discussed. In Section \ref{subsection23} methods relying on comparable corpora to build a bilingual lexicon are studied.

\subsection{Using Pivot languages}
\label{subsection21}
Over the past twenty years different approaches have been proposed to build a new source-pivot lexicon using a pivot language and consequently source-pivot and pivot-target dictionaries \citep{tanaka_construction_1994, istvan_bilingual_2009, tsunakawa_building_2008, tsunakawa_improving_2013, ahn_automatic_2006}. One of the most known and highly cited methods is the approach of Tanaka and Umemura \citep{tanaka_construction_1994} where they only use dictionaries to translate into and from a pivot language in order to generate a new dictionary. These pivot language-based methods rely on the idea that the lookup of a word in an uncommon language through a third intermediated language could be done with machines. Tanaka and Umemura use bidirectional source-pivot and pivot-target dictionaries (harmonized dictionaries). Correct translation pairs are selected by means of inverse consultation. This method relies on counting the number of pivot language definitions of the source word, which identifies the target language definition \citep{tanaka_construction_1994}. Sjobergh presented another well-known method in this field \citep{sjobergh_creating_2005}. He generated his English pivoted Swedish-Japanese dictionary where each Japanese-to-English description is compared with all Swedish-to-English descriptions. The scoring metric is based on word overlaps, weighted with inverse document frequency and consequently the best matches are selected as translation pairs. These two approaches \citep{tanaka_construction_1994, sjobergh_creating_2005} are the best-performing ones and are general enough to be applicable with other language pairs as well. The basis of most of the other ideas and approaches proposed in recent years is based on those two described approaches \citep{tanaka_construction_1994, sjobergh_creating_2005}.

Compared to other implementations, our approach needs a small and reliable extracted dictionary as a part of our seed input. The usage of this extracted dictionary is discussed in Section \ref{subsection32}. In our work, the \citep{sjobergh_creating_2005}'s method is used. Moreover, as we needed only the top translations with the highest scores the generality of a selected method was not a factor.
\subsection{Using Parallel Corpora}
\label{subsection22}
Another way to create a bilingual dictionary is to use parallel corpora. Using parallel corpora to find a word translation (i.e. word alignment) started with primitive methods of \citep{brown_statistical_1990} and continued with some other word alignment approaches such as \citep{gale_identifying_1991, gale_program_1993, melamed_portable_1997, ahrenberg_simple_1998, tiedemann_extraction_1998, och_improved_1999}. These approaches share a basic strategy of first having two parallel texts aligned in pair segments and second having word co-occurrences calculated based on that alignment. This approach usually reaches high score values of 90\% precision with 90\% recall, \citep{otero_learning_2007}. Many studies show that for well-formed parallel corpora high accuracy rates of up to 99\% can be achieved for both sentence and word alignment. Currently, almost the entire task of bilingual dictionary creation and especially the creation of a probability table for any word pairs could be done with well-known statistical machine translation software, GIZA++ \citep{och_systematic_2003}. Using Parallel corpora as the input of the dictionary creation process is attractive in two ways. First, alignment between sentences and words is very accurate as a natural characteristic of parallel corpora and these methods do not need any other external knowledge to build a bilingual lexicon. Second, no external bilingual dictionary (seed dictionary) is required. The main problem of creating a parallel corpus lexicon is the lack of extensive language pairs, therefore reliance on just using parallel corpora to build accurate bilingual dictionaries is impossible. For the selected languages in this work, Persian and Italian, the creation of an accurate bilingual dictionary based on the existing parallel corpora is not applicable, although using our low resource parallel corpora to create a small dictionary may be practical. This dictionary could be used as an input in other methods that would use this subordinate input to build a larger dictionary. This is used as the seed dictionary in approaches based on comparable corpora which are discussed in Section \ref{subsection32}.

\subsection{Using Comparable Corpora}
\label{subsection23}
There is a growing interest in the number of approaches focused on extracting word translations from comparable corpora \citep{fung_finding_1997, fung_ir_1998, rapp_automatic_1999, chiao_looking_2002, dejean_bilingual_2002, kaji_extracting_2005, otero_learning_2007, otero_automatic_2010, rapp_utilizing_2010, bouamor_building_2013, irimia_experimenting_2012, Morin2013, emmanuel2014looking}. Most of these approaches share a standard strategy based on context similarity. All of them are based on an assumption that there is a correlation between co-occurrence patterns in different languages \citep{rapp_identifying_1995}. For example, if the words “teacher” and “school” co-occur more often than expected by chance in a corpus of English, then the Italian translations of them, “insegnante” [teacher] and “scuola” [school] should also co-occur in a corpus of Italian more than expected by chance. The general strategy extracting bilingual lexicon from the comparable corpus could be described as follows:

\textit {Word target t is a candidate translation of word source s if the words with which word t co-occur within a particular window in the target corpus are translations of the words with which word s co-occurs within the same window in the source corpus.}

The goal is to find the target words having the most similar distributions with a given source word. The starting point of this strategy is a list of bilingual expressions that are used to build the context vectors of all words in both languages. This starting list is called the seed dictionary. The seed dictionary is usually provided by an external bilingual dictionary. \citep{dejean_bilingual_2002} uses one multilingual thesaurus as the starting list instead of using a bilingual dictionary. In \citep{otero_learning_2007} the starting list is provided by bilingual correlations previously extracted from a parallel corpus. In \citep{rapp_identifying_2012}, the authors extract a bilingual lexicon without using an existing starting list. Although they do not use any seed dictionary, their results are acceptable.

Another interesting issue considered in recent years is to evaluate the effect of the degree of comparability on the accuracy of extracted resources \citep{li_improving_2010,li_clustering_2011, Sharoff2013}. In \citep{li_improving_2010} the authors propose a metric calculating the degree of comparability and they use an iterative construction process to improve the quality of a given comparable corpus to extract a bilingual lexicon. In \citep{li_improving_2010} a cluster-based method is used to enhance corpus comparability based on the metric they introduced in \citep{li_clustering_2011} where it was claimed that most of the vocabulary of the initial corpus was preserved despite their changes in the corpus.

As described before, it is assumed that there is a small bilingual dictionary available at the beginning. Most methods use an existing dictionary \citep{rapp_automatic_1999, chiao_looking_2002, fung_finding_1997, fung_ir_1998} or build one with some small parallel resources \citep{otero_learning_2007}. Entries in the dictionary are used as an initial list of seed words. Texts in both source and target languages are lemmatized and part-of-speech (POS) tagged with function words are removed.

A fixed window size is chosen and it is determined how often a pair of words occurs within that text window. These windows are called the ``fixed-size window" which does not take into account word orders within a window. R. Rapp observed that word order of content words is often similar between languages, even between unrelated languages such as English and Chinese \citep{rapp_berechnung_1996}. In approaches considering word order, for each lemma, there is a context vector whose dimensions are the same as the starting dictionary but in different window positions with regard to that lemma. For instance, if the window size is 2, the first context vector of lemma A, where each entry belongs to a unique seed word, shows the number of co-occurrences two positions to the left of A for that seed word. Three other vectors should also be computed, counting co-occurrences between A and the seed words appearing one position to the left of A and the same for two right hand positions following lemma A. Finally, all four vectors of length $n$ are combined (where $n$ is the size of the seed lexicon) into a single vector of length $4n$. This method takes into consideration the word orders to define contexts. In this paper the efficiency of considering the word order schema is evaluated.

Simple context frequency and some additional weights such as inverse document frequency can be considered in bilingual lexicon construction approaches \citep{chiao_looking_2002}. Well-known and widely used weighting for these approaches is log-likelihood \citep{rapp_automatic_1999}. In this paper both frequencies, simple context and log-likelihood are evaluated and compared. In computation of the log-likelihood ratio, the simplified formula from Dunning and Rapp \citep{dunning_accurate_1993} is used:
\begin{equation}
\mathit{loglike}(A,B)=\sum _{i,j{\in}1,2}{{K}_{\mathit{ij}}\ast \log \frac{{K}_{\mathit{ij}}\ast N}{{C}_{i}\ast {R}_{j}}}
\end{equation}

Therefore:
\begin{multline}
\mathit{loglike}(A,B)=\\{K}_{11}\log \frac{{K}_{11}\ast N}{{C}_{1}\ast {R}_{1}}+ {K}_{12}\log \frac{{K}_{12}\ast N} {{C}_{1}\ast {R}_{2}}+ \\ {K}_{21}\log \frac{{K}_{21}\ast N}{{C}_{2}\ast {R}_{1}}+ {K}_{22}\log \frac{{K}_{22}\ast N}{{C}_{2}\ast {R}_{2}}
\end{multline}

Where:
\begin{equation}
{C}_{1}={K}_{11}+{K}_{12}
\end{equation}
\begin{equation}
{C}_{2}={K}_{21}+{K}_{22}
\end{equation}
\begin{equation}
{R}_{1}={K}_{11}+{K}_{21}
\end{equation}
\begin{equation}
{R}_{2}={K}_{12}+{K}_{22}
\end{equation}
\begin{equation}
N={C}_{1}+{C}_{2}+{R}_{1}+{R}_{2}
\end{equation}

With parameters \textit{K}\textit{\textsubscript{ij}} expressed in terms of corpus frequencies:

\textit{K}\textit{\textsubscript{11}} = frequency of common occurrence of word A and word \textit{B}

\textit{K}\textit{\textsubscript{12}} = corpus frequency of word A - \textit{K}\textit{\textsubscript{11}}

\textit{K}\textit{\textsubscript{21}} = corpus frequency of word B - \textit{K}\textit{\textsubscript{11}}

\textit{K}\textit{\textsubscript{22}} = size of corpus (no. of tokens) - corpus frequency of word \textit{A} - corpus frequency of word \textit{B}

All numbers have been normalized in our experiments.

For any word in a source language, the most similar word in a target language should be found. First, using a seed dictionary all known words in the co-occurrence vector are translated to the target language. Then, with consideration of the result vector, a similarity computation is performed to all vectors in the co-occurrence matrix of the target language. Finally, for each primary vector in the source language matrix, the similarity values are computed and the target words are ranked according to these values. It is expected that the best translation will be ranked first in the sorted list \citep{rapp_automatic_1999}. Different similarity scores have been used in the variants of the classical approach \citep{rapp_automatic_1999}. In \citep{laroche_revisiting_2010} the authors presented some experiments for different parameters like context, association measure, similarity measure, and seed lexicon. Some of the famous similarity metrics are included in the Appendix of this paper. We decided to use \verb"diceMin" similarity score in our work which has been used previously in \citep{curran_improvements_2002, plas_syntactic_2005,otero_learning_2007}. The \verb"diceMin" score is the the similarity of two vectors, X and Y is computed using below similarity measure.
\begin{equation}
diceMin(X,Y)= \frac{2\cdot\sum_{i=1}^n min(X_i,Y_i)}{\sum_{i=1}^n X_i+\sum_{i=1}^n Y_i}
\end{equation}

\section{Our Approach}
\label{section3}
Our experiments to build a Persian-Italian lexicon are based on the comparable corpora window approach discussed in Section \ref{subsection23}. An interesting challenge in our work is to combine different dictionaries with varying accuracies and use all of them as the seed dictionary for comparable corpora based lexicon generation. We address this problem using different strategies: First, combining dictionaries with some simple priority rules, and then, using all translations together with and without considering the differences in their weights. In Section \ref{subsection31}, our method to collect and create seed dictionaries and consequently, our implementation to use them independently is explained. In Section \ref{subsection32}, we describe the usage of comparable corpora to build a new Persian-Italian lexicon. Sections \ref{subsection33} and \ref{subsection34}, our approaches for combining three different dictionaries are explained. Section \ref{subsection35} described our proposed weighting method.
\subsection{Building Seed Dictionaries}
\label{subsection31}
We have used three different dictionaries and their combinations as the seed dictionaries. The first dictionary is a small Persian-Italian dictionary, the second dictionary is created based on the pivot-based method presented in \citep{sjobergh_creating_2005}, which contains top entries with the highest score, and the third dictionary is built using two little parallel Persian-Italian corpora. When there is more than one translation for an entry in the primary dictionary, we should select one translation. Most standard approaches select the first translation in the existing dictionary or the candidate with the highest score in the extracted (created) dictionary. However, in \citep{irimia_experimenting_2012}, several definitions for one word based on their scores could be selected in the seed dictionary generation step. Like other standard methods, we selected the first translation among all the candidates. In the following three subsections, our three dictionaries and the process of creating them are discussed.
\subsubsection{The Existing Dictionary – DicEx}
\label{subsubsection311}
We used a small Persian-Italian dictionary as the existing dictionary named \verb"DicEx". For each entry, only the first translation is selected to create lemmas. While \verb"DicEx" is a manually created dictionary, it is the most accurate dictionary in our experiments, and its size is the smallest in comparison with the other dictionaries.
\subsubsection{The Dictionary created by a Pivot based method – DicPi}
\label{subsubsection312}
We used the method introduced in \citep{sjobergh_creating_2005} as the baseline for the Pivot based dictionary creation. Translations with the highest scores are selected in this phase while removing the low scoring results. A Persian-English dictionary and an English-Italian dictionary are considered as inputs. All stop-words and all non-alphabet characters are removed from the English portion of these two dictionaries. Then the inverse document frequency, \textit{idf}, is calculated for the remaining English words as follows:
\begin{equation}
idf(w)=\log (\frac{|Pr|+|It|}{Pr_w+It_w})
\end{equation}
where $w$ is the word we calculate the weight for, ${\mid}$Pr${\mid}$ is the total number of dictionary entries in the Persian-English dictionary, ${\mid}$It${\mid}$ is the total number of dictionary entries in the English-Italian dictionary, Pr$_{w}$ is the number of descriptions in the Persian-English dictionary where the word $w$ occurs, and It$_{w}$ is this number for English-Italian.

In the next step, all the English descriptions in the first dictionary must be matched to all descriptions in the second. Matches are scored by word overlaps that are weighed by predefined inverse document frequencies. A word is only counted once regardless of the number of occurrences in the same description. Based on Sjobergh's method, \citep{sjobergh_creating_2005} scores are calculated as shown in Equation \ref{sjobergh_score}:
\begin{equation}
\label{eq:sjobergh_score}
score = \frac{2\cdot\sum_{w \in Pr \cap It} idf(w)}{\sum_{w \in Pr} idf(w) + \sum_{w \in It} idf(w)}
\end{equation}
where \textit{Pr} is the text in the translation part of Persian-English lexicon and \textit{It} is the translation text in the Italian-English Dictionary. When all scores are calculated, candidates with the highest score will be selected to build our new Persian-Italian dictionary. In contrast to the Sjobergh's implementation where the main focus is creating a dictionary with very large coverage, our goal is creating a small dictionary with more accuracy for use as a seed dictionary in the main system. Therefore, we select the top 40,000 translations from all translations and named it \verb"DicPi".

\subsubsection{The Dictionary extracted from Parallel corpora – DicPa}
\label{subsubsection313}
In this paper, we used our low parallel Persian-Italian resources (e.g. movie subtitles) to create a small dictionary by selecting the top translations with the highest probabilities. This parallel corpus-based dictionary, named \verb"DicPa", is used as the seed dictionary which is subsequently combined with other main dictionaries in the following phases. It is created from a general domain translation table automatically extracted with Giza++ \citep{och_systematic_2003}. When a word has more than one translation, only the highest probability translation is selected and others with lower probability are removed. Finally, we select the top entry words from the word-based translation table.
\subsection{Using seed dictionaries to extract lexicon from Comparable Corpora}
\label{subsection32}
Our window-based approach is presented in this section. Some mathematics and theoretical points of our approach were discussed in Section \ref{subsection23}. Given that there are large differences between Persian and Italian words in syntax and grammar, the window-based approach is preferred. The baseline of the method implemented in our study is an adaptation of \citep{rapp_automatic_1999}. Based on our proposed idea, the seed dictionary could be an existing dictionary, an automatically created dictionary, or a combination of them.

There are two types of input: the seed dictionary, and the bilingual comparable corpus. Weighting vectors must be created based on corpora and lexicons. Before the creation of matrices for both Persian and Italian languages, the stop words of corpora are deleted and it should be lemmatized. Two co-occurrence matrix sets are computed for the Persian and Italian corpora: one set for a simple approach and another one for ordered base approach. In order to check the effect of word orders in our experiments, we needed two matrices for our two corpora. These matrices have r rows where r is the number of unique words occurred in the corpus. If the size of our lexicon is $n$ and the selected size for windows is k, for the simple method which does not consider word orders, the matrices (for both Persian and Italian corpora) have $n$ columns where every column corresponds to a type word in the base lexicon. Each field ($i$,$j$), shows that how many times word $j$ is occurred with a distance of $k$ words, from word $i$ in the corresponding corpus. The size of this matrix is equal to $r \times n$.

In the order-based method, matrices must save the position of each word with pivot word in addition to saving the frequency in one window. We create it by dividing each field of the former matrix to $2n$ fields where each field shows a different position before or after pivot word where each new matrix is itself a $r \times n \times k$ matrix when the field ($i$,$j$,$k$) shows the number of times word $j$ has occurred in position indexed $k$ from word $i$.

Previous approaches show the need for replacing the co-occurrence frequency in the matrix by measures that are able to eliminate word-frequency effects and consequently to favor significant word pairs. Therefore we uses the log-likelihood ratio (i.e. Equation 1 \citep{dunning_accurate_1993}) in our approach described in Section \ref{subsection23}. To see its effect, we also carried out our tests without this metric by using the simple frequency matrix.

In order to calculate the similarity scores, we should transfer our matrices from the source language to the target language. The rows with all zero values are pruned from the Persian matrix. All remaining rows are considered as the potential translations. For each row, all columns are translated by using the seed lexicon; i.e. the source vectors are transferred to target vectors. Then, the similarity score for all possible translations is calculated. A possible translation is a row in the transferred matrix which corresponds to a row in the target matrix. Therefore the value of similarity scores should be calculated and sorted between any row in the transferred matrix and all the rows in the target matrix. In this experiment we use \verb"diceMin" similarity score described in Section \ref{subsection23} as the preferred score. In Section \ref{subsection35} of this paper, a new similarity score, \verb"newdiceMin" is proposed by the authors to weight dictionaries when different seed dictionaries are combined together.

In order to build a new lexicon, for each word (i.e. row) in the source vector, the best matches in the target vector could be considered as the translation. Therefore, for each entry, we select word corresponding to target vectors where the similarity score is more than the rest.
\subsection{Using simple combination}
\label{subsection33}
In this section, the process of creating the bigger seed dictionary by using a simple combination rule is discussed. The accuracy of the existed dictionary, \verb"DicEx" is highest among others and the accuracy of \verb"DicPi" is higher than the dictionary created from the parallel corpus (i.e. \verb"DicPa"). Based on the accuracy of dictionaries, a priority order is defined to create the final seed dictionary:

\begingroup
\centering
\verb"DicEx" ${>}$ \verb"DicPi" ${>}$ \verb"DicPa" \\
\endgroup

\noindent Our simple combination rule is:
\begin{adjustwidth}{+12pt}{+24pt}
Suppose that the priority of Dic$_{i}$ is more than the priority of Dic$_{j}$; if a word $w$ is in both Dic$_{i}$ and Dic$_{j}$, its translation is selected from Dic$_{i}$ (i.e. the dictionary with higher priority)\\
\end{adjustwidth}

\noindent By applying the above priority rule, a new Persian-Italian dictionary with more than 65,000 unique entries is created. We name this newly created dictionary \verb"DicCoSi". Apparently, all the words in \verb"DicEx" are included in \verb"DicCoSi". The experimental results show an improvement in the extracted lexicon when this new dictionary \verb"DicCoSi" is used as the main system's seed dictionary in comparison with using our three simple dictionaries individually.

\subsection{Using independent word combination}
\label{subsection34}
In our simple priority-based combination which is described in Section \ref{subsection33}, there is an important issue that should be discussed. Given two words, where the first one appears in all three dictionaries and the second one just appears in one dictionary. In our simple approach, there is no difference between these words. Therefore, a new advanced combination method is proposed. Our advanced combination method is based on the assumption that one word in two different dictionaries should be considered independently as two different words. For example, if a word appears in both dictionaries Dic$_{1}$ and Dic$_{2}$, it may have two independent columns in our vector matrix (i.e. it has two different weights in the transferred vectors). Therefore, the new dictionary named \verb"DicCoIn" is created where its size is equal to the sum of our three dictionary sizes. In this new dictionary, if the word $x$ occurs in two dictionaries, there are two different entries for it named x$_{i}$ and x$_{j}$ where $i$ and $j$ are the indicators of corresponding dictionaries.
\subsection{New weighting method}
\label{subsection35}
There is another problem in our proposed advanced combination. Even though some dictionaries are more accurate than others, there is no difference in dealing with initial seed dictionaries. In order to ease this problem, a new weighting model for similarity scores is introduced. This new metric relies on two following aspects:

(1) We could change the effect of each seed dictionary in order to consider higher weights for a more accurate dictionary. These weights could be tuned manually.

(2) If a word appears in two dictionaries, then it is not necessary to count it twice as a double-count would produce an unfair skew. We could consider its weight a little bit more than a normal occurrence weight and then divide it between different dictionaries.

If there are $k$ different dictionaries in our proposed independent word-based combination, to calculate the similarity scores between bilingual lemmas we could use the proposed equation \ref{eq:newdicemin}:
\begin{multline}
\label{eq:newdicemin}
newdiceMin(X,Y) = \\\frac{2\cdot\sum_{j=1}^k \sum_{X_i \in Dic_j} min(X_i,Y_i) \cdot w_j}{\sum_{i=1}^n X_i + \sum_{i=1}^n Y_i}
\end{multline}
where $n$ is the size of the new combined dictionary and w$_{j}$ is the weight of dictionary $j$. In our experiments, the size of $k$ is equal to three. This is apparent that if w$_{j}$ =1 for j=1, 2 and 3, then the method is the same as the previous approach described in Section \ref{subsection34}.
The new weighting method is based on this assumption that the dictionary with higher accuracy should affect the extracted lexicon more. In our experiments, two different sets of w$_{j}$ are studied and the results are evaluated in Section \ref{subsection53}.
\section{Preparing The Inputs}
\label{section4}
As stated prior, two primary inputs are needed to perform comparable corpora based lexicon generation: seed dictionary and comparable corpus/corpora. The procedures to prepare these needed data are described in sections 4.1 and 4.2,
To evaluate the result, a test dataset is needed. The evaluation of the test is performed by two people.

The first evaluator is one of the authors, who is a native Persian speaker and fluent in Italian and the second one is a Persian native who teaches the Italian language. If both of the evaluators agree on a translation word, it is accepted as a true translation, otherwise the translation is considered false. We selected 400 Persian objective test words from Nabid Persian-English dictionary \footnote{Nabid Dictionary, written by Hani Kaabi, Iran, 2002}. Since it is not appropriate to apply our approach for words that are already in the base lexicons, we removed all entries belonging to the 400 test words. The frequencies of all the selected words in our corpora (general corpus and specific domain corpus) were greater than 100.
\subsection{Seed Dictionaries}
\label{subsection41}
Three different seed dictionaries are used in our experiments. The first was a small preexisting Persian-Italian dictionary named \verb"DicEx". Another usage of this dictionary was to extract Persian-Italian parallel sentences from comparable corpora.

The second dictionary, \verb"DicPi", is a dictionary extracted by the pivot-based approach proposed in \citep{sjobergh_creating_2005}. To extract \verb"DicPi", two Persian-English and English-Italian dictionaries are needed. The Persian-English dictionary we used was the Nabid dictionary. It contains about 100,000 Persian index words. For the English-Italian dictionary, we used a personal dictionary created by the University of Pisa for their internal experiments\footnote{English-Italian words collected for projects DeSR Parser and Tanl Linguistic Pipeline, Prof. Giusseppe Attardi, Dipartimento di Informatica Università di Pisa}. This simple dictionary contains about 130,000 words. From the English portion of selected dictionaries, we removed the few stop words and all the characters that were not letters and then, the method described in Section \ref{subsection32} was applied to our data and a new dictionary was created. Although in the classic method \citep{sjobergh_creating_2005}, having a large coverage is very important; but in our experiment, we needed a smaller and more accurate extracted dictionary. Therefore, the top 40,000 words from all translation are extracted and inserted into the dictionary \verb"DicPi". We checked 200 randomly translated words and 84\% of them are acceptably translated. This accuracy is near but slightly less than the best results in the famous pivot based approaches described in Section \ref{subsection22}.

The parallel corpus-based lexicon, \verb"DicPa" was a word-to-word sub-part of a translation table, extracted with Giza++. Our parallel corpus contains about 29,000 sentences in both the Persian and Italian languages. These corpora are gathered from the Opus\footnote{OPUS, the open parallel corpus: http://opus.lingfil.uu.se/} database and WikiRetriever \footnote{Wikiretriever is a C++ crawler to retrieve very accurate parallel sentences from two different languages of Wikipedia website and has written to retrieve the limited amount of parallel sentences by using dates, numbers and famous words. The accuracy of this retriever for English and Persian, which has been evaluated before, is about 98\%, so is ideal for our purpose.}. When more than one definition was found for a word, the first one is selected and the others are discarded. Finally, the top 40,000 entries of the translation table with a high probability are selected as the new dictionary. Table \ref{tab:DictUsed} shows some characteristics of three dictionaries.

\begin{table}[h]
\begin{center}
\begin{tabular}{|c||c|c|}
\cline{1-3}
Dictionary Name & Entries & Mutual words\\ 
\cline{1-3}
\textit{DicEx} & 13,309 & N/A \\ 
\cline{1-3}
\textit{DicPi} & 40,000 & 6,954 \\ 
\cline{1-3}
\textit{DicPa} & 40,000 & 4,220 \\ 
\cline{1-3}
\end{tabular} 
\end{center}
\caption{Number of entries and mutual words with DicEx of dictionaries used in our Experiments}
\label{tab:DictUsed}
\end{table}

\subsection{Comparable Corpora}
\label{subsection42}
In our experiments, three different types of comparable corpora are gathered:
The first one is a small corpus of Wikipedia\footnote{Wikipedia, The Free Encyclopedia website: http://ww.wikipedia.org} articles in Persian and Italian extracted by WikiRetriever's preliminary phase \footnote{In the primary phase, Wikiretriever find articles which are about same issue to send it for next processes}. In order to skip those articles which are famous and well described in one of our languages (e.g. an article about an Italian village) we selected those article pairs where the difference between their sizes is not more than 50\%. After applying this criterion, 6,500 articles are selected in both languages: about 150,000 sentences for Persian and 176,000 sentences in Italian. Both groups of sentences were tokenized and lemmatized. The resulting corpus is called \verb"WikiCorpus" in our studies. This corpus is the most comparable corpus among our corpora (The comparability degree is more than the rest).

The second corpus is the international sport-related news gathered from different Persian and Italian news agencies. We used the ISNA\footnote{ISNA, Iranian students News Agency, International News part, Persian, http://isna.ir/fa/service/World} and the FARS \footnote{Fars News Agency, International News part, Persian, http://www.farsnews.com/newsv.php?srv=6} for the Persian part, and the news agency CORRIERE DELLA SERA\footnote{CORRIERE DELLA SERA, International news, Italian, http://www.repubblica.it/} and the Gazzetta dello Sport\footnote{La Gazzetta dello Sport, Italian, http://www.gazzetta.it/} for the Italian part. The numbers of selected articles are about 12,000 and about 15,000 from Persian and Italian resources, respectively. This corpus named SportCorpus, has more noise in comparison with Wikipedia created corpus but while international sports news are very similar in different agencies, the comparability degree is not too small. We combined SportCorpus and \verb"WikiCorpus" and used them together in our experimental results. We call this new combined corpus \verb"SpeCorpus" (Specific domain-based corpus).

The third corpus is based on international news gathered from different Persian and Italian news agencies. The difference between this corpus and \verb"SpeCorpus" is that the former was gathered from sport-related news and this one is gathered from general subjects. This is our biggest corpus but obviously has a very low comparability degree in comparison with \verb"SpeCorpus". The number of articles in the Persian version was about 108,000 and the Italian version contained about 140,000 articles. We used ISNA and FARS news agencies for the Persian version and CORRIERE DELLA SERA as the Italian resource. We named this corpus \verb"GenCorpus". By using \verb"GenCorpus" we could analyze the effect of using a very general corpus in comparison with specific domain-based corpus and we can see how the comparability degree of input corpus could affect the extracted lexicon.
\section{Experimental Results}
\label{section5}
As discussed in Section \ref{subsection32}, in order to see the effect of using order based windows, we considered both simple window and the word order based method, separately. The results show that taking orders into account is not very effective to extract Persian-Italian lexicons (i.e. only in a small number of cases, it has a slightly positive effect). The authors think the reason is the vast difference between the structures of Persian and Italian languages. However, in our experiments, we applied both schemas. Based on \citep{irimia_experimenting_2012}'s conclusion all window sizes were set to 5. In our approach, we have calculated both simple frequency and the log-likelihood ratio. Despite our expectation, in a few cases using simple co-occurrence has a more accurate result in comparison to using log-likelihood ratio. While this difference is very small, at most demonstrated figures in this paper, a simple frequency ratio is not considered and only log-likelihood ratio is shown.

All experiments described in this paper were applied on two types of comparable corpora: (1) the combination of \verb"WikiCorpus" and \verb"SportsCorpus" which we named \verb"SpeCorpus". (2) \verb"GenCorpus" as a big, general, and less comparable corpus. The characteristics of these corpora were discussed in Section \ref{subsection42}.

Finally, experiments are executed in order to evaluate our proposed combination models. In the first subsection, we use the three previously mentioned dictionaries as the individual seed lexicon. Then we used our two different proposed strategies to combine dictionaries and consequently the effect of the combinations is studied. Finally, in Section \ref{subsection53}, the new weighting model and its effect with different weight sets are evaluated.

In our experiments and for each test, two different result sets are calculated. The Top-1 measure is the number of times when the test word's acceptable translation is ranked first, divided by the number of test words. The Top-10 measure is equal to the number of times a correct translation for a word appears in the top 10 translations in the resulting lexicon, divided by the number of test words. The evaluation process is performed manually by two evaluators. As described in Section \ref{section4}, one translation is assumed to be true if both evaluators agree.
\subsection{Using independent dictionaries}
\label{subsection51}
\begin{figure*}[h]
\centering
\includegraphics[width=4in]{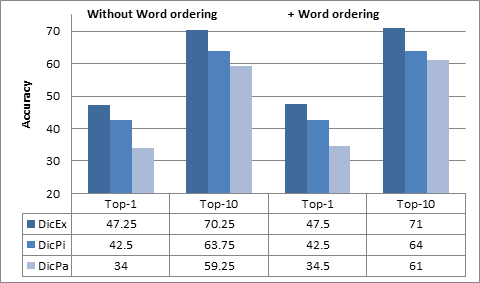}
\caption{Results of using independent dictionaries with and without considering word orders. All results are based on log-likelihood measurement using SpeCorpus (in-domain corpus)}
\label{Fig1}
\end{figure*}

\begin{figure*}[h]
\centering
\includegraphics[width=4in]{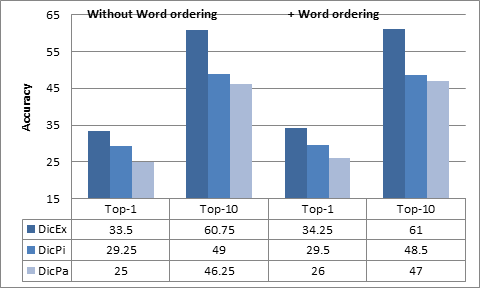}
\caption{Results of using independent dictionaries with and without considering word orders. All results are based on log-likelihood measurement using GenCorpus (general corpus)}
\label{Fig2}
\end{figure*}

\begin{figure*}[h]
\centering
\includegraphics[width=4in]{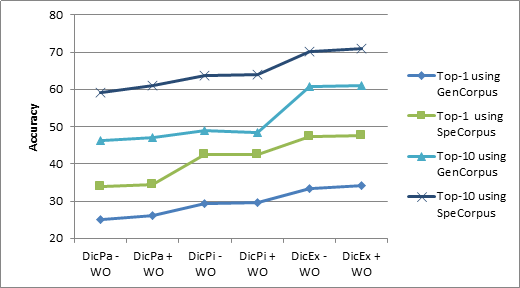}
\caption{Effect of using different corpora in with different comparability degree}
\label{Fig3}
\end{figure*}

In the first phase of our experiments, all three previously mentioned dictionaries are used individually as the seed lexicon. These are the preexisting dictionary (\verb"DicEx"), the pivot base extracted dictionary (\verb"DicPi") and the parallel corpus-based dictionary (\verb"DicPa").
Figures \ref{Fig1} and \ref{Fig2} summarize the evaluation results using these three seed dictionaries with and without using word order. The goal of this experiment is to see the effect of using different comparable corpora.
Figure \ref{Fig1} shows the results of using \verb"SpeCorpus", a corpus with a higher comparability degree and Figure \ref{Fig2} demonstrates the results of using \verb"GenCorpus", the bigger corpus with lower comparability degree. A comparison between these two corpora and the effects of using them individually is illustrated in Figure \ref{Fig3}. The goal of this classification is to determine the effect of using comparable corpora with a higher comparability degree. Figure \ref{Fig3} shows that using a corpus with a higher comparability degree increases the accuracy in both Top-1 and Top-10 results significantly. As it is expected, this difference for Top-1 results is more than the Top-10 measure. However, it should be understood that our test words were selected from those words having a frequency more than our threshold. When \verb"SpeCorpus" is much smaller than \verb"GenCorpus", this may increase the chance of finding a better translation using it. For the rest of the experiments in this paper, \verb"SpeCorpus" is used as a comparable corpus and \verb"GenCorpus" is ignored.

Another option shown in Figures \ref{Fig1} and \ref{Fig2}, is the effect of considering word orders in lemma vectors. As we expected, this could increase efficiency slightly. However, as discussed prior, and because of the vast differences between the structure of Italian and Persian languages, this improvement is very small and could be negligible as a conclusion.
According to results, and our expectation, the \verb"DicEx" has better outcome despite its small size compared with the other dictionaries. A reason is the high accuracy of \verb"DicEx" as it is a handmade dictionary. We could consider it a 100\% accurate dictionary. The experimental results show that \verb"DicPi" has a slightly better efficiency in comparison with parallel corpora based dictionary \verb"DicPa". The authors conclude that the reason is the limitation of our parallel Persian-Italian corpus used to create a translation table. Therefore we have selected some unimportant words that are not translation decisive. If we had enough parallel corpora, this conclusion could be supported or rejected strongly. However based on retrieved statistics in section 4.1 (Table \ref{tab:DictUsed}), \verb"DicPi" has more mutual words with preexisting dictionary (\verb"DicEx") in comparison with \verb"DicPa"'s mutual words with \verb"DicEx". This issue could be used to predict the accuracy order in our work.

\begin{figure*}[h]
\centering
\includegraphics[width=4in]{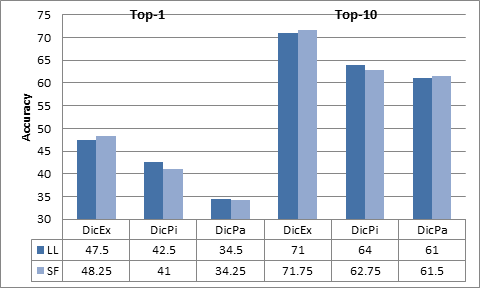}
\caption{The effect of log-likelihood with using SpeCorpus}
\label{Fig4}
\end{figure*}

In Figure \ref{Fig4}, the effect of using log-likelihood in comparison with using the simple frequency vectors is shown. For each experiment, we used two different schemas: with and without considering word orders. Before this experiment we expected that using log-likelihood and word ordering together has the best efficiency but based on the results, when the seed dictionary is \verb"DicEx" or \verb"DicPa", simple frequency schema has a slightly better accuracy in comparison with the log-likelihood ratio even though this superiority is very narrow and could be neglected. For \verb"DicPi" both schemas have almost the same outcome (i.e. in one test case using log-likelihood has better result and in other one using the simple frequency). In the legend of Figure \ref{Fig4}, \verb"SF" and \verb"LL" means using simple frequency and log-likelihood respectively. In general and with considering the noise effects, this hypothesis could be supported that based on our data sets, none of these schemas has a better efficiency in comparison with other.

\subsection{Using composite dictionaries}
\label{subsection52}
In this section, we evaluate our ideas of combining different dictionaries together. As described before, two different types of combination are used in our experiments. The simple combination creates a dictionary using a simple priority rule and the advanced combination for all dictionaries considering all translations of any word. Table \ref{tab:DictCombo} shows the results of these studies. According to this table, the best results for Top-1 measure belong to the simple combination model when all dictionaries are combined together. The best Top-10 results belong to the advanced combination model combining all dictionaries. In advanced combination, all words in all dictionaries are selected in the lexicon generation phase, and this generally gives us better Top-10 results. An important issue for our advanced combination is that all translations in different dictionaries have the same weight and this may decrease the effect of \verb"DicEx". Although it is our most accurate dictionary, it is also the smallest one. This problem is tackled in the next section by using our weighting lemma.\\

\begin{table*}[h]
\begin{center}
\begin{tabular}{|c r|r|r|r|}
\cline{1-5}
\multirow{2}{*}{Dictionary Name} & \multicolumn{2}{|c|}{Top-1} & \multicolumn{2}{c|}{Top-10} \\ 
\cline{2-5}
& \multicolumn{1}{|c|}{Simple} & Advanced & Simple & Advanced \\
\cline{1-5}
\multirow{1}{*}{DicEx + DicPi} & \multicolumn{1}{|r|}{50.00} & 49.50 & 75.00 & 75.50 \\
\cline{1-5}
\multirow{1}{*}{DicEx + DicPa} & \multicolumn{1}{|r|}{48.75} & 48.00 & 74.00 & 74.75 \\
\cline{1-5}
\multirow{1}{*}{DicPi + DicPa} & \multicolumn{1}{|r|}{42.50} & 43.00 & 66.75 & 67.50 \\
\cline{1-5}
\multirow{1}{*}{All Dictionaries\footnote{Named \textit{DicCoSi} for simple combination and \textit{DicCoIn} for advanced combination.}} & \multicolumn{1}{|r|}{\textit{50.25}} & 49.75 & 75.25 & \textit{76.75} \\
\cline{1-5}
\end{tabular}
\end{center}
\caption{The effect of different dictionaries in combination with different methods on \textsc{SpeCorpus} for advanced combination}
\label{tab:DictCombo}
\end{table*}

\subsection{Using new weighting}
\label{subsection53}
In this section, we describe the proposed weighting method to use different dictionaries together. The goal of introducing this metric is to "tune" the impact of a dictionary considering its accuracy and correctness. Two different heuristics are considered to adjust weights in this part. The first one is to tune weights based on dictionaries accuracy. The accuracies could be collected from Top-10 scores calculated in phase 5.1 (results in the second column of Figure \ref{Fig1}) . In the first set, the weights for \verb"DicEx", \verb"DicPi" and \verb"DicPa" are 0.7, 0.64 and 0.59, respectively.

In our second heuristic set, the weights are calculated based on both accuracy and the dictionary size. This weight set is constructed based on the assumption that the bigger dictionary should have a lower effect on the final result. We used the following formula to calculate the weights.

\begin{equation}
w_i = accuracy_i \cdot \frac{MaxSize}{size_i}
\end{equation}

Therefore, based on the second heuristic, Equation 12, and with considering the results in our study the weights are:

\vspace{8pt}
\begingroup
\centering
W$_{\verb"DicEx"}$ = 2.10,\\W$_{\verb"DicPi"}$ = 0.64,\\W$_{\verb"DicPa"}$ = 0.59.\\
\endgroup
\vspace{12pt}
The results of these experiments based on different weighting sets are shown in Table \ref{tab:WeightSchemaEffect}. W$_{i}$ = 1 presents the classical approach without using the proposed weighting system.

\begin{table}[h]
\centering
\begin{tabular}{l r r|}
\cline{2-3}
& \multicolumn{1}{|c|}{Top-1} & \multicolumn{1}{c|}{Top-10} \\
\cline{1-3}
\multicolumn{1}{|l}{W$_i$=1\footnote{Same as the previous results (without any priority)}} & \multicolumn{1}{|r|}{50.25} & 76.75 \\
\cline{1-3}
\multicolumn{1}{|l}{Weight 1} & \multicolumn{1}{|r|}{52.50} & 78.25 \\
\cline{1-3}
\multicolumn{1}{|l}{Weight 2} & \multicolumn{1}{|r|}{\textbf{53.75}} &	\textbf{81.25} \\
\cline{1-3}
\end{tabular}
\caption{The effect of new weighting schema on the accuracy of the extracted dictionary (In all tests, the combination of three dictionaries is used and the comparable corpus is \textsc{SpeCorpus})}
\label{tab:WeightSchemaEffect}
\end{table}

Table \ref{tab:WeightSchemaEffect} shows that when we consider the accuracy of dictionaries, the extracted lexicon has better accuracy in comparison with the advanced combination. The best efficiency belongs to the second weighting set where we consider both accuracy and size together when the weight of the most accurate dictionary, \verb"DicEx" is much higher than the rest.

\begin{figure*}[h]
\centering
\includegraphics[width=4in]{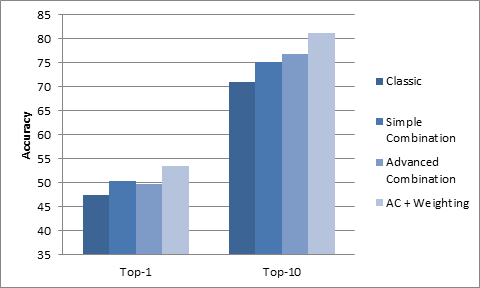}
\caption{The effect of different introduced combinations}
\label{Fig5}
\end{figure*}

Finally, Figure \ref{Fig5} shows a brief illustration to see the effect of our combination methods in comparison with classic approaches when they used just the existing dictionary, \verb"DicEx" (the most accurate independent dictionary in our study) as the seed dictionary. In all results, the log-likelihood ratio with considering word ordering issue is used to extract bilingual lexicons from \verb"SpeCorpus", our corpus with high comparability degree. In legends of this figure, \verb"AC" means advanced combination model.\\

\section{Conclusion}
\label{section6}
In the last decade, some methods have been proposed to extract bilingual lexicons from comparable corpora. In order to create a Persian-Italian lexicon, we decided to implement a comparable corpora-based lexicon generation method. This type of methods usually need a small dictionary as their starting seed dictionary. In our study, three different seed lexicons (and combinations) are used consisting of one preexisting dictionary and two extracted dictionaries. The first extracted dictionary is based on parallel-corpora dictionary creation methods and the second one is extracted by pivot language models. While for a seed dictionary a small dictionary is needed, we just selected the top translations from these created dictionaries. In the first part of our study, the effects of using these dictionaries on different types of comparable corpora are evaluated.

A new and interesting challenge was introduced in our work combining different dictionaries creating the seed dictionary. We used two different strategies: First, composing dictionaries with some priority rules; second, using all dictionaries together considering similar words in two dictionaries as a different word. Both of these strategies were studied and based on our experimental results these novel dictionary combinations could improve the efficiency of the results. The proposed advanced dictionary combination is almost as accurate as our simple combination. Furthermore, in all experiments, the effect of the comparability degree of the initial comparable corpus is studied using different types of comparable corpora. The results show that a higher degree of comparability in input corpus, has a more accurate lexicon despite the fact that the less comparable corpus is larger in comparison with the higher comparable corpus; although using a specific corpus may decrease the generality of the extracted lexicon.

Finally, a new weighting method has been proposed to increase the efficiency of our dictionary combination. This new weighting method uses the assumption that the effect of a more accurate seed dictionary should have a better result in comparison with others; the experimental result shows that using a more accurate weighting system causes the extracted lexicon to be more accurate.

\section*{Acknowledgments}
The research was partially supported by OP RDE project No. 
CZ.02.2.69/0.0/0.0/16\_027/0008495, International Mobility of Researchers at 
Charles University. The authors gratefully acknowledge the contribution and help of Dr. Fatemeh Alimardani, Dr. Daniele Sartiano, Vahid Pooya, Amir Onsori, S. M. H. Mirsadeghi, and Dr. Mahshid Nikravesh to this work.

\bibliography{acl2017}
\bibliographystyle{acl_natbib}

\appendix
\section*{Appendix}
Different similarity scores have been used in the variants of the classical approach of extracting bilingual lexicon from comparable corpora; \citep{rapp_automatic_1999} used city-block as their preferred similarity vector. The cosine similarity is used by \citep{fung_finding_1997, fung_ir_1998, chiao_looking_2002, Saralegui_2008} and the lin similarity metric is used by \citep{lin_98short}. The other well-known similarity metrics are dice and jaccard \citep{chiao_looking_2002}. In both dice and jaccard metrics, the association values of two lemmas with the same context are joined using their product. There are two different forms of jaccard and dice; the jaccardMin metric \citep{grefenstette_explorations_1994, kaji_extracting_1996} and \verb"diceMin" \citep{curran_improvements_2002, plas_syntactic_2005, otero_learning_2007}. Only the smallest association weight is considered for both of these lemmas. The jaccard and the dice are very similar based on results gathered in \citep{Otero_2008} which authors discuss the efficiency of several similarity metrics combined with simple occurrences and log-likelihood weighting schemes. In \citep{laroche_revisiting_2010} the authors presented some experiments for different parameters like context, association measure, similarity measure, and seed lexicon. In recent works, the similarity of two vectors, X and Y is computed using one of these similarity measures:

\begin{flushleft}
\begin{equation}
\mathit{city}_{\mathit{block}}(X,Y)=\sum _{i=1}^{n}{|{X}_{i}-{Y}_{i}|}
\end{equation}
\end{flushleft}

\begin{equation}
\mathit{cosine}(X,Y)=\frac{\sum _{i=1}^{n}{({X}_{i}.{Y}_{i})}}{\sqrt{\sum _{i=1}^{n}{{{X}_{i}}^{2}}}\sqrt{\sum _{i=1}^{n}{{{Y}_{i}}^{2}}}}
\end{equation}
\begin{equation}
\mathit{diceMin}(X,Y)=\frac{2\ast \sum _{i=1}^{n}{\mathit{min}({X}_{i},{Y}_{i})}}{\sum _{i=1}^{n}{{X}_{i}}+\sum _{i=1}^{n}{{Y}_{i}}}
\end{equation}

\begin{multline}
\mathit{diceProd}(X,Y)=\\\frac{2\ast \sum _{i=1}^{n}{({X}_{i}.{Y}_{i})}}{\sum _{i=1}^{n}{{{X}_{i}}^{2}}+\sum _{i=1}^{n}{{{Y}_{i}}^{2}}}
\end{multline}

\begin{equation}
\mathit{jaccardMin}(X,Y)=\frac{\sum _{i=1}^{n}{\mathit{min}({X}_{i},{Y}_{i})}}{\sum _{i=1}^{n}{\mathit{max}({X}_{i},{Y}_{i})}}
\end{equation}

\begin{multline}
\mathit{jaccardProd}(X,Y)=\\\frac{\sum _{i=1}^{n}{({X}_{i}.{Y}_{i})}}{\sum _{i=1}^{n}{{{X}_{i}}^{2}}+\sum _{i=1}^{n}{{{Y}_{i}}^{2}}-\sum _{i=1}^{n}{({X}_{i}.{Y}_{i})}}
\end{multline}

\begin{equation}
\mathit{lin}(X,Y)=\frac{\sum _{{X}_{i},{Y}_{i}{\neq}0}{({X}_{i}+{Y}_{i})}}{\sum _{i=1}^{n}{{X}_{i}}+\sum _{i=1}^{n}{{Y}_{i}}}
\end{equation}

\end{document}